\documentclass[runningheads]{llncs}
\usepackage{graphicx}
\usepackage{booktabs}
\usepackage{float}
\usepackage{gensymb}
\usepackage{tabularx}
\usepackage{tikz}
\usepackage{comment}
\usepackage{amsmath,amssymb} 
\usepackage{color}
\usepackage{amsfonts}
\usepackage{caption}
\usepackage{subfigure}
\usepackage{multirow}
\usepackage{mathtools}
\usepackage{tablefootnote}

\usepackage{tikz}
\usepackage{comment}
\usepackage{amsmath,amssymb} 
\usepackage{color}
\usepackage[pagebackref,breaklinks,colorlinks,citecolor=darkgreen]{hyperref}

\usepackage[accsupp]{axessibility}  

\usepackage[width=122mm,left=12mm,paperwidth=146mm,height=193mm,top=12mm,paperheight=217mm]{geometry}

\newcommand{\OURS}{PCR-CG}

\definecolor{Gray}{gray}{0.92}
\definecolor{darkgreen}{rgb}{0.13, 0.55, 0.13}
\newsavebox\CBox
\def\textBF#1{\sbox\CBox{#1}\resizebox{\wd\CBox}{\ht\CBox}{\textbf{#1}}}

\begin{document}
\pagestyle{headings}
\mainmatter
\def\ECCVSubNumber{6316}  

\title{\OURS: Point Cloud Registration via Deep Explicit Color and Geometry}

%
\author{Yu Zhang\inst{1} \and
Junle Yu\inst{2} \and
Xiaolin Huang\inst{1} \and
Wenhui Zhou\inst{2} \and
Ji Hou\inst{3}
}
%
%
\institute{Shanghai Jiaotong University, Shanghai, China \and
Hangzhou Dianzi University, Hangzhou, China \and
Technical University of Munich, Munich, Germany
}
\maketitle
\begin{center}
\includegraphics[width=1.0\linewidth]{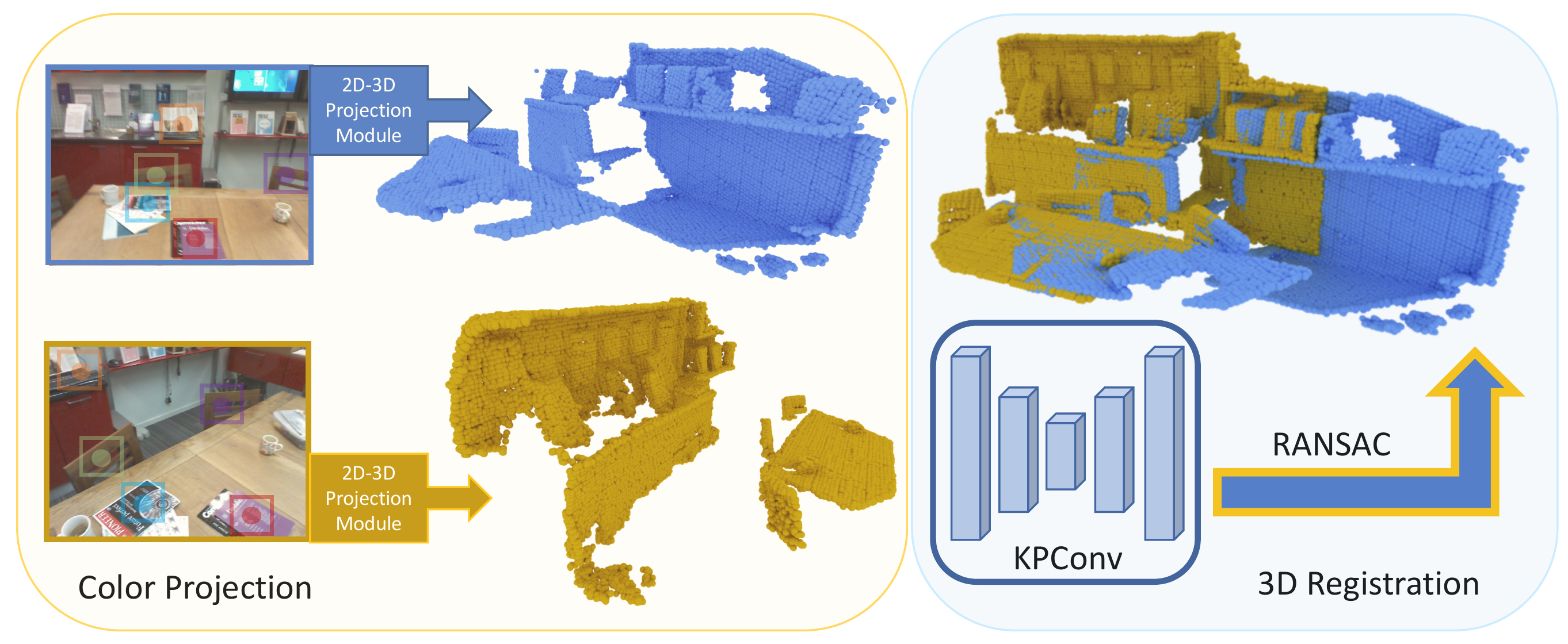}
\captionof{figure}{
We seek to align two point clouds in RGB-D data. To better leverage RGB information, we propose \OURS{}, a 2D-3D projection module that explicitly lifts 2D deep color features to 3D geometry representation. A pair of RGB-D frames are used as input, where each RGB-D frame is composed of a color image and a depth frame. 3D geometry is represented by the point cloud that is generated from depth frame. We leverage a pre-trained 2D network to predict correspondences between frames and extract regional features from color images. The 2D regional features are further lifted to 3D via our proposed 2D-3D projection module in an explicit manner. Our code is open sourced at: \url{https://github.com/Gardlin/PCR-CG}}
\label{fig:teaser}
\end{center}

\begin{abstract}
In this paper, we introduce \OURS: a novel 3D point cloud registration module explicitly embedding the color signals into the geometry representation. Different from previous methods that only use geometry representation, our module is specifically designed to effectively correlate color into geometry for the point cloud registration task. Our key contribution is a 2D-3D cross-modality learning algorithm that embeds the deep features learned from color signals to the geometry representation. With our designed 2D-3D projection module, the pixel features in a square region centered at correspondences perceived from images are effectively correlated with point clouds. In this way, the overlapped regions can be inferred not only from point cloud but also from the texture appearances. Adding color is non-trivial. We compare against a variety of baselines designed for adding color to 3D, such as exhaustively adding per-pixel features or RGB values in an implicit manner. We leverage Predator~\cite{Huang_2021_CVPR} as the baseline method and incorporate our proposed module onto it. To validate the effectiveness of 2D features, we ablate different 2D pre-trained networks and show a positive correlation between the pre-trained weights and the task performance. Our experimental results indicate a significant improvement of $6.5\%$ registration recall over the baseline method on the 3DLoMatch benchmark. We additionally evaluate our approach on SOTA methods and observe consistent improvements, such as an improvement of $2.4\%$ registration recall over GeoTransformer as well as $3.5\%$ over CoFiNet. Our study reveals a significant advantages of correlating explicit deep color features to the point cloud in the registration task.
\end{abstract}

\section{Introduction}
With commodity depth sensors commonly available, such as Kinect series, a variety of RGB-D datasets are created~\cite{dai2017scannet,zeng20173dmatch,armeni_cvpr16,song2014sliding}. With recent breakthroughs in deep learning and the increasing prominence of RGB-D data, the computer vision community has made tremendous progress on analyzing point cloud~\cite{qi2017pointnet} and images~\cite{he2016deep,he2017mask}. Recently, we observe a rapid progress in cross modality learning between geometry and colors~\cite{hou2021pri3d,liu2020p4contrast,liu2021contrastive,srinivasan2017learning,chang2017matterport3d,balntas2017pose}. However, they are mainly focused on high-level semantic scene understanding tasks, such as semantic/instance segmentation~\cite{dai20183dmv,lahoud20193d} and object detection~\cite{imvotenet}. Compared to high-level tasks, cross-modality learning between color and geometry is less explored in low-level tasks, such as point cloud registration. In this paper, we discuss correlating RGB priors for aligning two partial point clouds.

Point cloud registration has been fast developed because of its wide applications~\cite{qin2022geometric,Huang_2021_CVPR,aoki2019pointnetlk,yu2021cofinet,el2021unsupervisedr,bai2021pointdsc}; its 2D counter-part has been developed even earlier and achieved great success~\cite{niethammer2019metric} in many systems, such as visual SLAM~\cite{stuckler2014combining}. Mainstream methods adopt first-correspondences-then-transformation manner, namely estimating transformations between two frames based on the correspondence matching. In this context, correspondence-matching-based methods~\cite{sarlin20superglue,zhou2021patch2pix} have showed appealing results in the 2D domain. However, current methods in 3D merely use geometry as the only input. Therefore, exploring to combine RGB is demanded and of great importance to the point cloud registration task. In this manner, a variety of existing 2D approaches and pre-trained models can also be further leveraged in 3D point cloud registration.

Finding correspondences is essential for calculating the transformation matrix between two frames, and correspondences only appear in the overlap region. In this context, estimating the overlap regions of two frames is critical for point cloud registration. Intuitively, we can tell the overlap regions not only from geometric inputs like point cloud, but also from color signals like images. With this observation, we propose to embed color signals into point cloud representation, so as to effectively predict 3D correspondences for the registration task. To this end, we propose \OURS{}, a novel module that explicitly embeds RGB priors into the geometry representation for the point cloud registration.

In our work, we build upon successful Predator~\cite{Huang_2021_CVPR}, following the standard point cloud registration pipeline, namely first finding correspondences and then using RANSAC to estimate the rotation and translation matrices between two frames of point clouds. To enable the usage of RGB values from captured RGB-D data, our approach introduces three steps. First, a 2D pre-trained neural network~\cite{sarlin20superglue} is used to predict 2D correspondences between pure RGB frames. Based on the correspondences, we extract square regions centered at each correspondence pixel. Furthermore, another 2D pre-trained neural network summarizes the features from pixels in each region. We investigate the effectiveness of 2D pre-trained features in the 3D task by trying different 2D pre-trained backbones, such as ImageNet and Pri3D~\cite{hou2021pri3d} pre-trained models. We note that the 2D models are pre-trained on different datasets. In this context, the transfer ability of 2D part shows promising results. In this manner, we are able to take advantage of massive existing 2D pre-trained models. Secondly, we propose a 2D-3D projection module to explicitly project the 2D features to 3D point cloud region by region, according to the camera intrinsic and transformation matrix. We exhaustively explore the possible designs, e.g., implicitly concatenating per-pixel features to points. And we demonstrate that the design of explicitly projecting overlap-aware regions surpass the rest in our ablation studies.

Following Predator~\cite{Huang_2021_CVPR}, we evaluate our work on 3DMatch and the more competitive and difficult 3DLoMatch~\cite{Huang_2021_CVPR} benchmark. For a fair comparison, we use the same 3D backbone as Predator. In both benchmarks, we observe significant improvements in our proposed color and geometry learning strategy. Our approach outperforms the state-of-the-art method by a large margin of $6.5\%$ registration recall on the 3DLoMatch benchmark. \\

In summary, the contributions of our work are four-fold:
\begin{itemize} \itemsep0em 
    \item We introduce a novel 2D-3D projection module that explicitly embeds the 2D color into the point cloud for registration task. 
    \vspace{0.2cm}
    \item We experimentally show that our method outperforms the baseline~\cite{Huang_2021_CVPR} by a significant gap of $6.5\%$ registration recall on the challenging 3DLoMatch benchmark.
    \vspace{0.2cm}
    \item Our approach is agnostic to backbones and brings consistent improvements to SOTA methods, such as +$2.4\%$ registration recall over GeoTransformer~\cite{qin2022geometric} and +$3.5\%$ over CoFiNet~\cite{yu2021cofinet}. 
    \vspace{0.2cm}
    \item We conduct empirical studies and show the transfer ability of 2D pre-trained weights for 3D point cloud registration tasks.
\end{itemize}

\section{Related Work}
Advances in deep learning enable fast development in many high-level and low-level tasks. In this section, we firstly review point cloud and image registration tasks, and then discuss a few additional relevant works in the area of multi-modal learning across color and geometry. \\

\noindent \textbf{Point Cloud Registration.} 
Point Cloud Registration plays an important role in computer vision community. Most successful methods in this field start with a low-level task, namely correspondence matching. A transformation matrix can be then estimated from the predicted correspondences. Finding correspondences have been researched even before deep learning era. Traditional machine learning methods and hand-crafted descriptors, such as ICP~\cite{arun1987least,besl1992method} and SIFT~\cite{lowe2004distinctive}, have drawn great attention back then. And the field is moving even faster since deep learning era. Leveraging the powerful deep learning features to learn rotation-invariant descriptors~\cite{choy2019fully,bai2020d3feat,yu2022riga} for correspondences that are further fed into RANSAC for registration is the most successful story nowadays. Following the same pipeline, Predator~\cite{Huang_2021_CVPR} achieves the state-of-the-art results and first proposes to solve the registration problem on low-overlap frames. However, they only use geometry as the single-source input. In this paper, we build upon their framework and propose an effective module that explicitly fuses the overlap regions learned from 2D color signals. \\

\noindent \textbf{Image Registration.} 
As the counterpart of 3D point cloud registration, 2D image registration also contributes significantly to the computer vision community. It enables many high-level applications, such as 3D Reconstruction and visual SLAM~\cite{schoenberger2016sfm,schoenberger2016mvs}. Compared to 3D point cloud registration, 2D image registration uses only color input~\cite{sarlin2020superglue,revaud2019r2d2} and takes advantage of many existing pre-trained 2D network, such as ResNet with ImageNet pre-trained weights. The success of 2D image registration shows the possibility of learning registration from pixel input. Besides, the motivation of taking advantage of massive existing 2D pre-trained models suggests incorporating 2D signal into 3D registration task. In this work, we explore how to effectively use the 2D signal on 3D registration task. \\

\noindent \textbf{2D-3D Multi-Modality Learning.}
Joint learning from color and geometry signals has been researched in many high-level tasks, such as in both 2D and 3D scene understanding~\cite{hou20193d,dai20183dmv,hou2020revealnet,imvotenet,hou2021pri3d,liu2020p4contrast,hu2021bidirectional,liu20213d}. 3D-SIS~\cite{hou20193d} proposes to implicitly leverage the color signal for 3D instance segmentation and detection tasks. RevalNet~\cite{hou2020revealnet} adopts the similar idea of implicitly fusing color and geometry for 3D instance completion task. Panoptic3D~\cite{dahnert2021panoptic} infers geometry and 3D panotpic segmentation from pure 2D inputs. ImVoteNet~\cite{imvotenet} adds a 2D detector in addition to VoteNet~\cite{qi2019deep} to explicitly use 2D color input. 3D-to-2D Distillation~\cite{liu20213d} presents a method to fuse 3D features for 2D semantic segmentation tasks. BPNet~\cite{hu2021bidirectional} uses a bidirectional projection module to mutually learn 2D-3D signals for both 2D and 3D semantic segmentation tasks. Besides scene understanding tasks, 2D-3D learning is also explored in representation learning. Pri3D~\cite{hou2021pri3d} proposes to learn 2D representation in a pre-training paradigm for 2D scene understanding. P4Contrast~\cite{liu2020p4contrast} learns 3D representation from a novel 2D-3D loss for 3D scene understanding. However, most of the previous research focus on high-level semantic tasks. In this paper, we discuss the color-geometry learning in the point cloud registration task focusing on the low-level domain, i.e., predicting correspondences for point cloud registration. Additionally, we study the transfer ability of 2D network in the 3D registration task.

\section{Methods}
\label{method}
In this section, we introduce our proposed method \OURS{}. Our method builds upon Predator~\cite{Huang_2021_CVPR}, and additionally introduces the color input besides geometry via our 2D-3D projection module. In Section~\ref{data_representation}, we introduce the data representation of both color and geometry. In Section~\ref{sec:projection_module}, we detail our module specifically designed for lifting color inputs into 3D. At last, we introduce our training strategy as well as the setup of transferred 2D network for extracting features from color inputs.

\begin{figure}[t!]
\begin{center}
\includegraphics[width=1.0\linewidth]{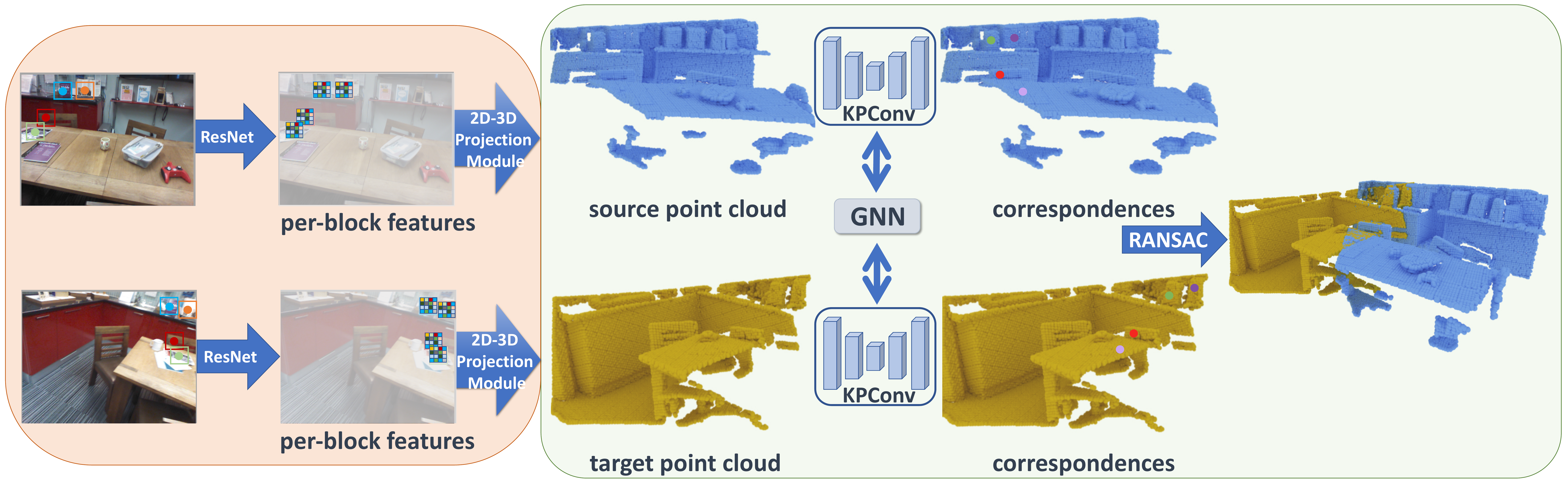}
\end{center}
\vspace{-0.25cm}
   \caption{\textbf{\OURS{} Pipeline.} The pipeline is composed of a 3D network, a 2D network and a 2D-3D projection module. Both 3D geometry and 2D images are taken as input and used to jointly learn features for detecting correspondences. The 2D network takes RGB images as input and extracts per-region features. A 2D-3D Projection Module is used to lift 2D pixel features into 3D point cloud. The concatenated features are fed into 3D network for finding correspondences. Due to our 2D-3D projection module, the 3D supervision can pass gradients back to the 2D network, and, therefore, yield an end-to-end training. }
\label{fig:network}
\vspace{-0.2cm}
\end{figure}

\subsection{Data Representation}
\label{data_representation}
Color and geometry data have different representations. In our method, we use point cloud to represent the geometry input similar to Predator. At the same time, the RGB images are used as the input to our proposed module.\\

\noindent \textbf{Geometry Data.} Each training sample contains a pair of non-aligned RGB-D frames. The transformation matrix that aligns them is used as the ground truth. We use point cloud lifted by depth frame as geometric input and predict the correspondences between them. For pre-computing the ground truth of correspondences, we transform one point cloud to the other, according to the aforementioned transformation matrix. Then, correspondences are found by a nearest neighbor search within a threshold in the Euclidean space.\\

\noindent \textbf{Color Data.} We evaluate our method mainly in RGB-D datasets, namely 3DMatch and 3DLoMatch. In these datasets, each point cloud is fused by 50 consecutive depth frames. The RGB images and depth images are aligned in these RGB-D datasets. Therefore, each point cloud is also associated with 50 RGB frames. We pick up the first and the last RGB images in the 50 consecutive frames for training and validating our 2D-3D projection module. Each RGB image is resized to the resolution of 240x320 in pixels. Notably, we do not need ground truth for 2D data, the 2D network is pre-trained on other data. 

\subsection{Projection Module}
\label{sec:projection_module}
\noindent \textbf{Insertion of the Projection Module.}
Before entering into our method, we first revisit Predator. In Predator, a point cloud is input to a 3D neural network. In the encoder, attention modules are used to correlate features obtained from source and target frames. The correlated features are fed into a decoder. The final layer outputs a score for each point to indicate its likelihood on overlapped regions. Per-point features from the final layer are used for finding correspondences. Next, correspondences are ranked based on the scores, and topk correspondences’ features are fed into RANSAC. Finally, RANSAC consumes the features of selected correspondences to further estimate the transformation matrix between the source and target frames. In this context, our module is directly inserted at the beginning of the 3D network without interfering the rest of the pipeline. The overview of the pipeline is showed in Figure~\ref{fig:network}. \\

\begin{figure}[h!]
\begin{center}
\includegraphics[width=0.9\linewidth]{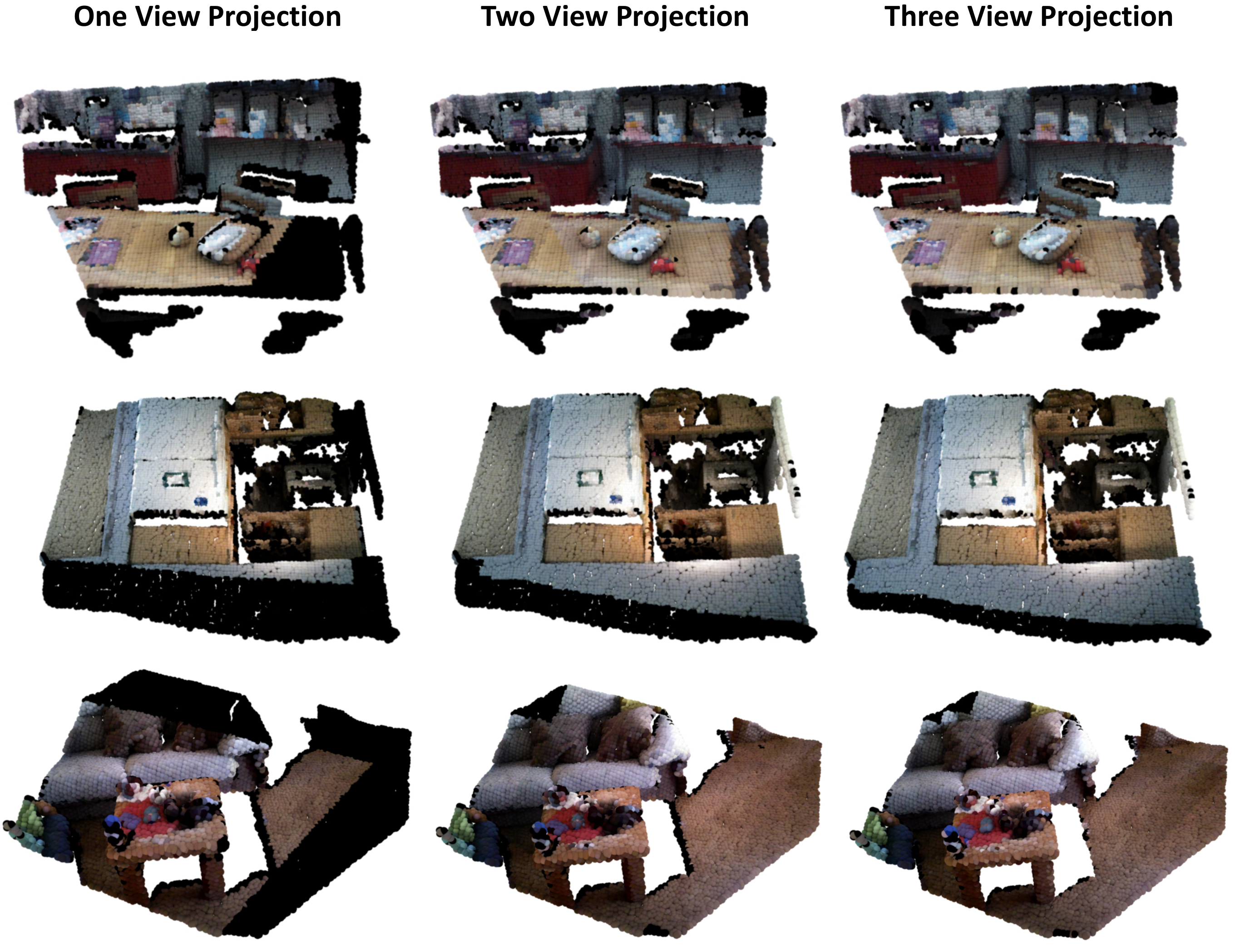}
\end{center}
   \caption{\textbf{Color Coverage Ablation.} Non-black points could take 2D features; black points indicate no features taken from 2D. We observe that with one image, not every point can be associated with color features; with two views, most points have the coverage of projected color features. However, adding the third view does not significantly improve the cover coverage. Therefore, our approach takes two views as the default setup. }
\label{fig:color_ablation}
\end{figure}

\begin{figure}[tp]
\begin{center}
\includegraphics[width=0.95\linewidth]{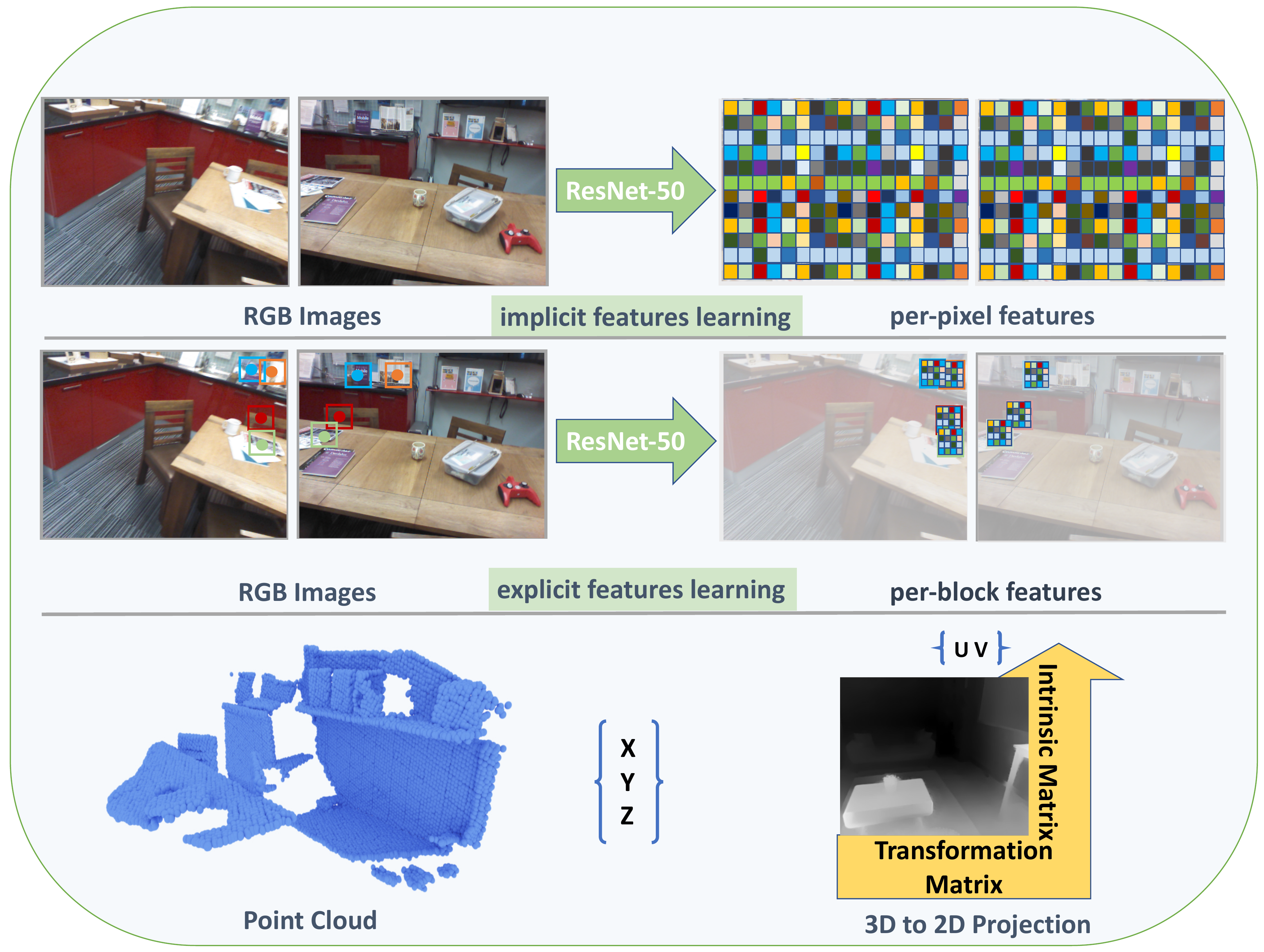}
\end{center}
   \caption{\textbf{2D-3D Projection Module.} We introduce a novel 2D-3D projection module to lift deep color features into 3D. The module takes the transformation matrix and depth map to project regional features to point cloud.}
\label{fig:project_module}
\end{figure}

\noindent \textbf{Lifting 2D to 3D.} To train on both color and geometry inputs, we propose a novel module that embeds deep color features into 3D representation. Our module~\OURS{} takes RGB images as input and converts them to features that are lifted to 3D. More specifically, the input RGB images to the module are corresponding to the input point clouds. The 3D network consumes a pair of point clouds, while our 2D-3D projection module takes a pair of RGB images. To concatenate the features of 2D pixels into 3D points, we project its xyz coordinates of each point cloud onto its associated image planes. In our setup, we select the first and the last RBG images among 50 RGB-D frames that are used to generate the point cloud. Since each point cloud is tied to two color images, we average the feature vectors sampled from the overlapped regions. In the end, we can append the feature vectors from 2D pixels to 3D points. We illustrate this projection procedure in Figure~\ref{fig:project_module}. The 3D network remains the same as Predator~\cite{Huang_2021_CVPR}. The only difference is that we adjust input dimensions of the first layer, as we concatenate the features from 2D counter part. The combined features are fed into the KPConv encoder and are crossed at bottleneck part via attention modules, which is the same as Predator~\cite{Huang_2021_CVPR}.

\noindent \textbf{Pre-trained 2D Networks.} We empirically find that simply appending RGB values to 3D points does not significantly improve the performance. Similar results are observed in ImVoteNet~\cite{imvotenet} and 3DMV~\cite{dai20183dmv}, and we confirm this observation in the low-level task as well. Therefore, we propose to lift deep color features rather than RGB values to 3D. In our newly designed module, the 2D network for extracting pixel features is a standard ResUNet-50 backbone. We choose ResUNet since its encoder weights can be initialized by most popular 2D pre-trained models, such as ImageNet, Pri3D~\cite{hou2021pri3d} and SuperGlue~\cite{sarlin20superglue}. In this manner, we can easily leverage various existing pre-trained models. In the ablation study, we indicate a significant influence of different 2D pre-trained weights on the 3D registration results. \\

\noindent \textbf{Implicit vs. Explicit Projection.} We introduce additionally a 2D network for extracting color features. However, it is a debate whether to use the color implicitly or explicitly. Implicit projection lifts the feature vector of each pixel to 3D, while the other one leverages the 2D information explicitly. More specifically, we use a pre-trained 2D estimator, i.e., SuperGlue~\cite{sarlin20superglue}, to predict the correspondences between the color images of source and target frames. Then, we project features extracted from the regions around the correspondences. In this manner, the regions lifted explicitly to 3D indicate a rough overlap estimated from color signals, which significantly improve the 3D correspondence matching results. We experimentally demonstrate the advantages of explicit project over implicit one in the ablation study. \\

\noindent \textbf{Frame Selection.} Each partial point cloud is fused by 50 consecutive depth frames. We propose to select the first and the last frames considering the trade-off between performance and efficiency. Regarding the number of selected frames, we present the color coverage in Figure~\ref{fig:color_ablation}. We show an increasing number of registration recall with more views in the ablation study.

\section{Experimental Results}
\label{results}

In this section, we show experimental results and ablation studies regarding our proposed 2D-3D projection module. We mainly focus on indoor data, specifically \textit{3DMatch} and \textit{3DLoMatch} benchmarks. For the main results, we show Registration Recall (RR), Feature Matching Recall (FMR) and Relative Rotation Error (RRE), as well as Relative Translation Error (RTE). Different number of points is sampled for registration in the main result. Additionally, we conduct ablation studies on the influences of 2D backbones pre-trained on different datasets, influences on color coverage in point cloud, and also end-to-end training in Section~\ref{sec:ablation}. \\

\noindent \textbf{Experiments Setup.} For training, we use an SGD optimizer with learning rate 0.005 and a batch size of 1. We use exponential learning scheduler, and the
learning rate is decreased by a factor of 0.95 every epoch. The model is trained on a single GPU (RTX3090) for 30 epochs. In test, we use open3D implementation for feature matching with nearest neighbor searching and RANSAC estimation. For 2D Network, we use ResUNet-50 with a ResNet-50 encoder to extract per-pixel features. We download the ImageNet pre-trained model, SuperGlue and Pri3D pre-trained models from official sources. In all experiments, KPConv~\cite{thomas2019kpconv} is used as 3D Backbone for extracting geometric features from point cloud. \\

\noindent \textbf{Metrics.} We mainly compare the results on four metrics, namely Registration Recall (RR), Feature Matching Recall (FMR), Relative Rotation Error (RRE), and Relative Translation Error (RTE). RR is the main metric we compare on and is most reliable, representing the fraction of pairs of point cloud, for which the correct transformation parameters are found after correspondence matching and RANSAC. Similar to Predator, we also report Feature Match Recall (FMR), defined as the fraction of pairs that have at least 5\% inlier matches with less than 10 cm residual under the ground truth transformation, even if the transformation cannot be recovered from those matches. Relative Translation (RTE) and Rotation Errors (RRE) measure the deviations from the ground truth pose. More specifically, the RTE is computed by the differences between two frames by L1 norm; RRE is the drifted degrees between two frames registered by predicted transformation. Please refer to supplementary materials for detailed explanations and mathematical definitions. \\

\begin{table}[h]
\centering
\begin{tabular}{l|ccccc|ccccc}
\toprule
& \multicolumn{5}{c}{\textBF{3DMatch}}&  \multicolumn{5}{c}{\textBF{3DLoMatch}}\\
\# Sampled Points~  &~5000~&~2500~&~1000~&~500~&~250~~~&~~5000~&~2500~&~1000~&~500~&250~\\
\midrule
\midrule
&\multicolumn{10}{c}{\textit{Feature Matching Recall}(\%) $\uparrow$}  \\
\midrule
3DSN~\cite{gojcic2019perfect} &~95.0~&~94.3~&~92.9~&~90.1~&~82.9~~~&~~63.6~&~61.7~&~53.6~&~45.2~&~34.2~\\
FCGF~\cite{choy2019fully} &~97.4~&~\underline{97.3}~&~\underline{97.0}~&~96.7~&~\underline{96.6}~~~&~~76.6~&~75.4~&~74.2~&~71.7~&~67.3~\\
D3Feat~\cite{bai2020d3feat} &~95.6~&~95.4~&~94.5~&~94.1~&~93.1~~~&~~67.3~&~66.7~&~67.0~&~66.7~&~66.5~\\
SpinNet~\cite{ao2021spinnet} &~\underline{97.4}~&~97.0~&~96.4~&~\underline{96.7}~&~94.8~~~&~~75.5~&~75.1~&~74.2~&~69.0~&~62.7~\\
Predator~\cite{Huang_2021_CVPR} &~96.6~&~96.6~&~96.5~&~96.3~&~96.5~~~&~~\underline{78.6}~&~\underline{77.4}~&~\underline{76.3}~&~\underline{75.7}~&~\underline{75.3}~\\
\midrule
Ours -- \OURS{} &~\textbf{97.4}~&~\textBF{97.5}~&~\textBF{97.7}~&~\textBF{97.3}~&~\textBF{97.6}~~~&~~\textBF{80.4}~&~\textBF{82.2}~&~\textBF{82.6}~&~\textBF{83.2}~&~\textBF{82.8}~\\
\midrule
&\multicolumn{10}{c}{\textit{Registration Recall}(\%) $\uparrow$} \\
\midrule
3DSN~\cite{gojcic2019perfect} &~78.4~&~76.2~&~71.4~&~67.6~&~50.8~~~&~~33.0~&~29.0~&~23.3~&~17.0~&~11.0~\\
FCGF~\cite{choy2019fully}  &~85.1~&~84.7~&~83.3~&~81.6~&~71.4~~~&~~40.1~&~41.7~&~38.2~&~35.4~&~26.8~\\
D3Feat~\cite{bai2020d3feat}  &~81.6~&~84.5~&~83.4~&~82.4~&~77.9~~~&~~37.2~&~42.7~&~46.9~&~43.8~&~39.1~\\
SpinNet~\cite{ao2021spinnet}  &~88.8~&~88.0~&~84.5~&~79.0~&~69.2~~~&~~58.2~&~56.7~&~49.8~&~41.0~&~26.7~\\
Predator~\cite{Huang_2021_CVPR}&~\underline{89.0}~&~\underline{89.9}~&~\textBF{90.6}~&~\underline{88.5}~&~\underline{86.6}~~~&~~\underline{59.8}~&~\underline{61.2}~&~\underline{62.4}~&~\underline{60.8}~&~\underline{58.1}~\\
\midrule
Ours -- \OURS{} &~\textBF{89.4}~&~\textBF{90.7}~&~\underline{90.0}~&~\textBF{88.7}~&~\textBF{86.8}~~~&~~\textBF{66.3}~&~\textBF{67.2}~&~\textBF{69.0}~&~\textBF{68.5}~&~\textBF{65.0}~\\
\bottomrule
\end{tabular}
\vspace{0.3cm}
\caption{Results on \emph{3DMatch} and \emph{3DLoMatch}. Our holistic approach combining explicit deep color and geometric features results in significantly improved results over previous approaches including the most recent CoFiNet. \OURS{} surpasses our baseline Predator~\cite{Huang_2021_CVPR} by a large margin. Note that our approach uses the same backbone and pipeline as Predator and does not include the coarse-to-fine technique introduced by CoFiNet~\cite{yu2021cofinet}. Pri3D pre-trained model and two-view projection (explicit) are used for our approach.}
\vspace{-0.35cm}
\label{tab:main_results}
\end{table} 

\begin{figure}[t!]
\begin{center}
\includegraphics[width=1.0\linewidth]{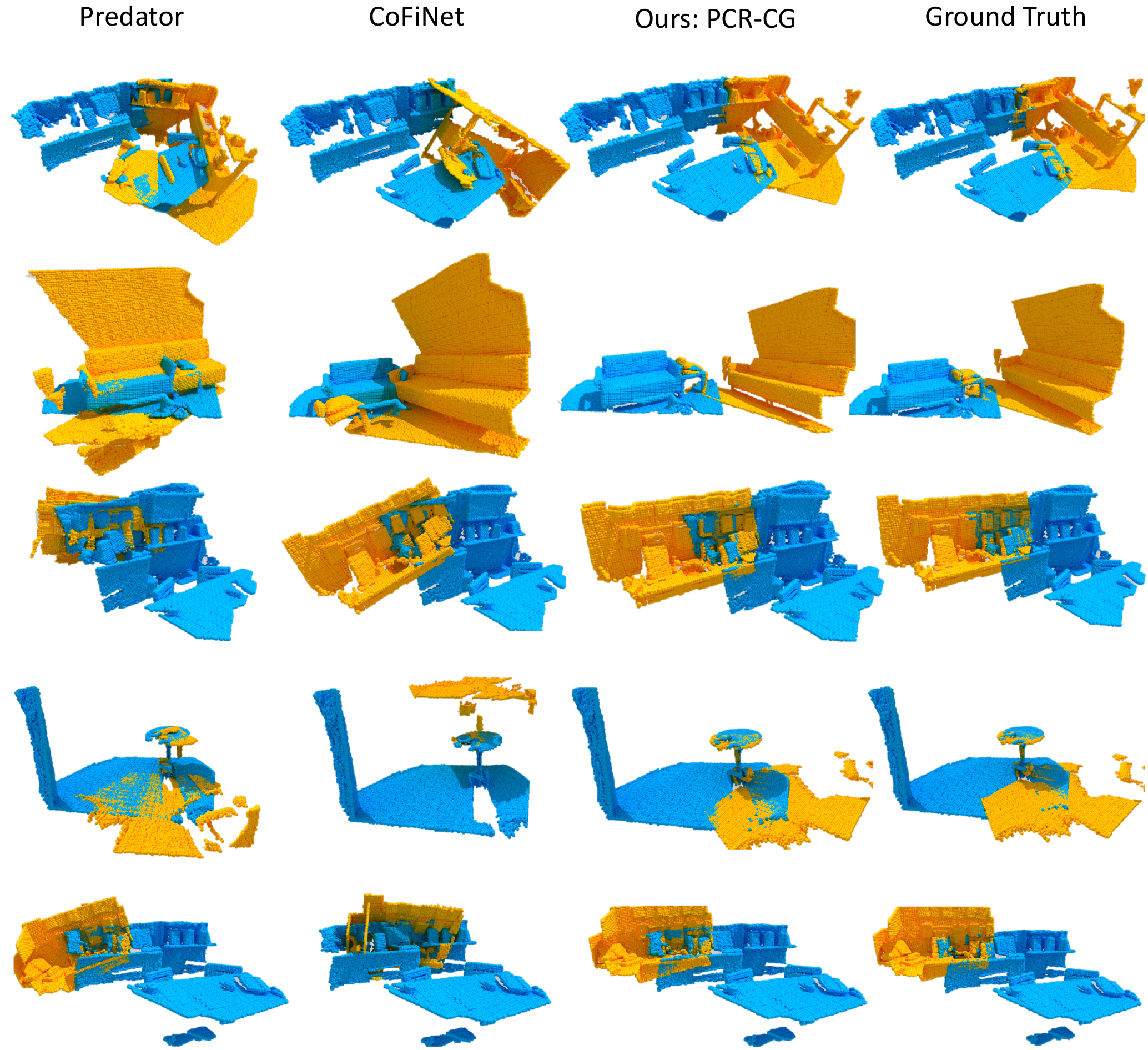}
\end{center}
   \caption{\textbf{Qualitative Comparisons on 3DLoMatch.} We demonstrate visual comparisons between \OURS{} and other SOTA methods. With the help of \OURS{}, the registration results are much more accurate.}
\label{fig:qualitative_compare}
\vspace{-0.5cm}
\end{figure}

\noindent \textbf{3DLoMatch Benchmark.} We show the results on 3DLoMatch in Table.~\ref{tab:main_results}. In 3DLoMatch, each pair of frames has at most $30\%$ overlaps, and therefore it is a more challenging and difficult benchmark. We compare our methods with state-of-the-art methods in terms of Registration Recall and Feature Matching Recall. We show that our method outperforms previous SOTA algorithms by a large margin on different numbers of sampled points, which indicates the effectiveness of explicitly embedding the deep color features into 3D. More specifically, we surpasses our baseline Predator by a significant margin especially with less sampled points, e.g., +7.7\% on RR and +7.5\% on FRM respective with 500 sampled points. In Table~\ref{tab:update_rre_rte}, we demonstrate that our approach also outperforms previous methods on Relative Rotation and Translation Errors. Besides the quantitative results, we also show qualitative results in 3DLoMatch benchmark in Figure~\ref{fig:qualitative_compare}. 

\noindent \textbf{3DMatch Benchmark.} We additionally report numbers on 3DMatch, where the overlap between two partial frames is at least 30\%. Compared to 3DLoMatch, 3DMatch is an easier and almost saturated benchmark. In Table~\ref{tab:main_results}, our method outperforms the baseline method Predator in both Registration Recall (RR) and Feature Matching Recall (FMR) in most cases, except for 1000 sampled points in Registration Recall (RR), where our method achieves the second best performance. In Table~\ref{tab:update_rre_rte}, our method also achieves state-of-the-art results on Relative Rotation Error (RRE) and Relative Translation Error (RTE).

\begin{table}[h]
    \centering
    \begin{tabular}{l|cc|cc}\toprule
         &\multicolumn{2}{c}{\textbf{3DMatch}}  & \multicolumn{2}{c}{\textbf{3DLoMatch}} \\ 
         &~~\textit{RRE} (\degree)~~&~~\textit{RTE} (m)~~&~~\textit{RRE} (\degree)~~&~~\textit{RTE} (m)~~\\ 
    \midrule
    3DSN~\cite{gojcic2019perfect}         & 2.199 & 0.071  & 3.528 & 0.103 \\
    FCGF~\cite{choy2019fully}       & \textBF{1.949} & 0.066  & 3.146 & 0.100    \\
    D3Feat~\cite{bai2020d3feat}          & 2.161& 0.067    & 3.361 & 0.103   \\
    Predator~\cite{Huang_2021_CVPR}        & 2.029 & \underline{0.064}  & \underline{3.048} & {0.093}   \\
    CoFiNet~\cite{yu2021cofinet}         & 2.002 & \underline{0.064}  & {3.271} & \underline{0.090}   \\  \midrule
    Ours -- \OURS{}  & \underline{1.993} & \textBF{0.061} & \textBF{3.002} & \textBF{0.087}  \\
     \bottomrule
    \end{tabular}
    \vspace{0.25cm}
    \caption{\textbf{Additional Quantitative Results.} We further show Relative Rotation and Translation Errors with 5,000 sampled points on 3DMatch and 3DLoMatch benchmarks. Our approach outperforms previous methods by a large margin and achieves the best in RTE on both benchmarks as well as the best RRE on 3DLoMatch and the second best in RRE on 3DMatch. The improved results suggest the effectiveness of our method in point cloud registration task.}
    \label{tab:update_rre_rte}
    \vspace{-0.5cm}
\end{table}

\subsection{Ablation Study}
\label{sec:ablation}
In this section, we ablate different designs of projecting 2D to 3D for point cloud registration task. We show our design of explicitly leveraging the deep color features achieves the best result. Furthermore, we show the significant influence of the 2D network and the frame selection on the final 3D registration results. We conduct our ablation experiments in \textit{3DLoMatch} benchmark.

\noindent \textbf{Frame Selection and Color Coverage.} In \textit{3DLoMatch}, each point cloud is fused by 50 consecutive frames. To ensure $100\%$ color coverage, 50 frames must be all used for each point cloud. However, 50 times forward and backward passes are time-consuming. To ensure the most color as well as efficiency, we propose to use the first and last frame to back-project pixel features into 3D geometry. The visuals in Figure~\ref{fig:color_ablation} demonstrate that there are approximately $30\%$ points that are not covered with one frame. With two views, most points are covered. With three views, it only slightly improves the color coverage. Quantitatively, the registration recall confirms the observation. To provide in-depth analysis of how much influence it has in terms of number of views, we adjust the numbers of images used in the training and testing procedure. In Table~\ref{tab:color_ablation}, we can clearly see the signals that using more views leads to a better registration recall, which proves that the proposed projection module contributes significantly to the 3D registration task. \\

\begin{figure*}[t]
\begin{center}
\includegraphics[width=1.0\linewidth]{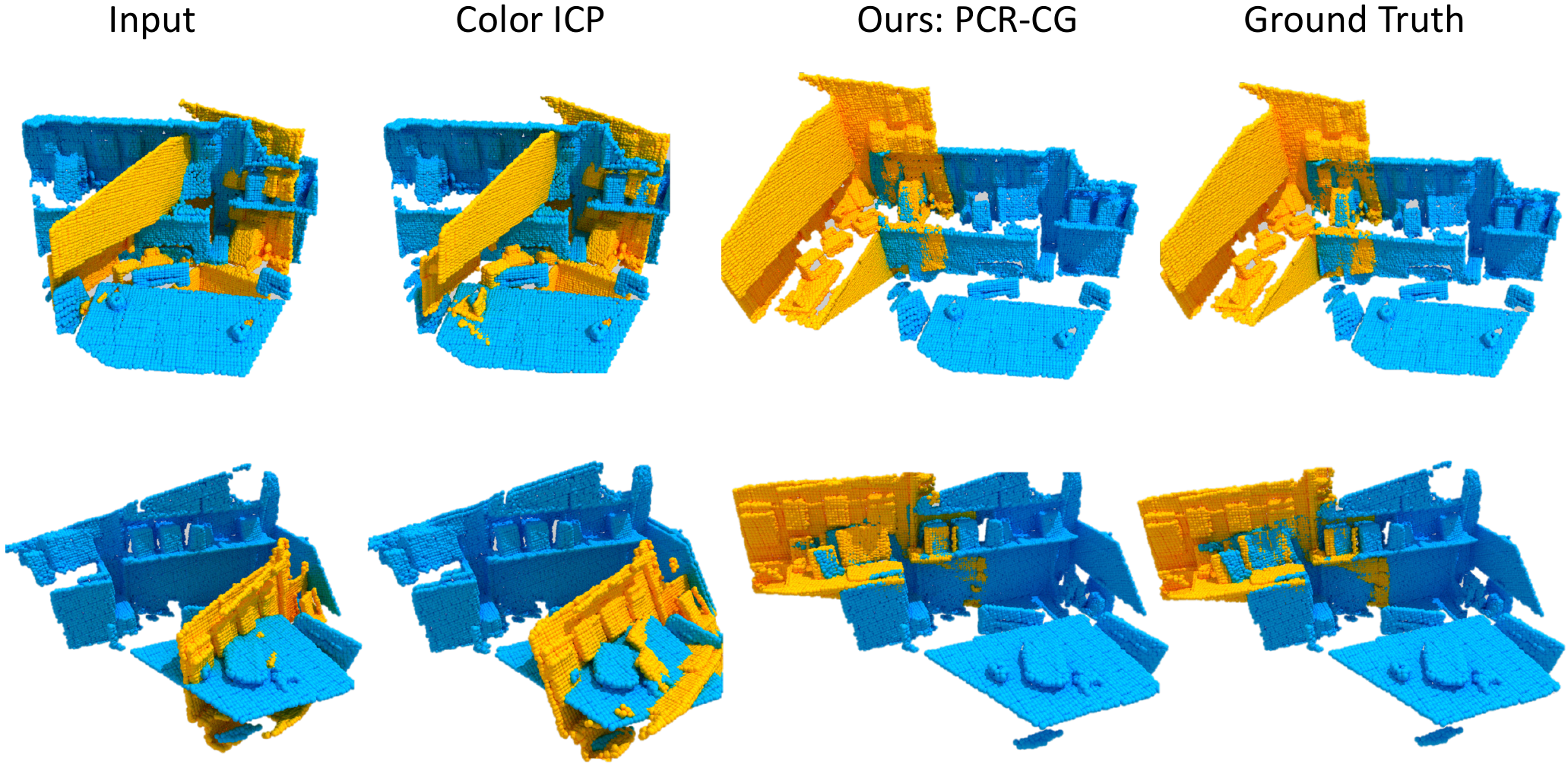}
\end{center}
   \caption{\textbf{Qualitative Comparisons with RGBs-based Methods}. Compared to ColorICP~\cite{park2017colored}, our method is more robust to initialization, especially in the case of large transformation and low overlaps such as in 3DLoMatch.}
\label{fig:color_icp}
\vspace{-0.35cm}
\end{figure*}

\noindent \textbf{Deep Color Features vs. RGBs.} We compare our method to ColorICP~\cite{park2017colored} in Tab.~\ref{tab:color_icp} to show the effectiveness of deep color features. Similar to ICP, ColorICP also requires a good pose initialization. With original pose initialization that has a large transformation, both ICP and ColorICP failed to align two point clouds (see Fig.~\ref{fig:color_icp}). With improved poses estimated by our method as initialization, both show marginal improvements. This observation demonstrates the importance of our method by embedding deep rather than shallow color features into geometry. Similarly, we use SIFT estimated from RGB images to register their point clouds. This result indicates the same conclusion and is showed in Table~\ref{tab:IC:sfm_pose_3dlomatch}. \\

\begin{table}[h]
    \centering

    \begin{tabular}{l|cc|cc}\toprule
         &\multicolumn{2}{c}{\textit{Original Initial Pose}} & \multicolumn{2}{c}{\textit{Improved Initial Pose}} \\ 
            &~~3DMatch~~&~~3DLoMatch~~&~~3DMatch~~&~~3DLoMatch~\\ 
            \midrule
            ICP~\cite{rusinkiewicz2001efficient}~ &   4.20 & 1.40 & 91.0 & 68.5 \\             
            ColorICP~\cite{park2017colored}~  &   4.90 & 1.50 & \textBF{91.4} & \textBF{68.8} \\
            OURS -- \OURS{}~~             &   \textBF{90.7} & \textBF{68.2} & -- & -- \\
            \bottomrule
    \end{tabular}
    \vspace{0.25cm}
    \caption{\textbf{Registration Recall on Initialzations}. Our method outperforms ICP and ColorICP on both \textit{3DMatch} and \textit{3DLoMatch} benchmarks by large margins and more robust to the bad pose initialization. Improved Initial Pose refers to using \OURS{}-estimated poses as an initialization to ICP and ColorICP. }
    \label{tab:color_icp}
    \vspace{-0.5cm}
\end{table}

\begin{table}[h]
    \centering
    \begin{tabular}{l|cc}\toprule
                  &~~3DLoMatch~~&~~3DMatch~~  \\ \midrule
                SIFT-DLT~~~ &0.4 & 0.9\\
     \bottomrule
    \end{tabular}
    \vspace{0.25cm}
    \caption{We utilize OpenCV SIFT-DLT~\cite{lowe2004distinctive} to calculate relative image pose and corresponding registration recall on 3DLoMatch benchmark.}
    \label{tab:IC:sfm_pose_3dlomatch}
\end{table}

\noindent \textbf{Window size.} We ablate different window sizes for extracting deep color features, and empirically find window size 11x11 achieves the best performance (see Table~\ref{tab:window_size_ablation}).\\

\begin{table}[h]
    \centering
    \begin{tabular}{l|c|ccccc}\toprule
&~\textit{Window size}~   & \multicolumn{5}{c}{\textit{Registration Recall (\%)}} \\ 
         &     &~~5000~~&~~2500~~&~~1000~~&~~500~~&~~250~~\\ 
    \midrule
     \OURS{}~~  &   3x3  &~~63.1~~&~~64.0~~&~~65.1~~&~~64.6~~&~~60.7~~\\
     \OURS{}~~  &   7x7  &~~64.7~~&~~66.4~~&~~67.1~~&~~65.8~~&~~62.6~~\\
     \OURS{}~~  &  11x11 &~~\textBF{66.3}~~&~~\textBF{67.2}~~&~~\textBF{69.0}~~&~~\textBF{68.5}~~&~~\textBF{65.0}~~\\
     \OURS{}~~ &   17x17  &~~64.1~~&~~65.7~~&~~66.2~~&~~65.6~~&~~62.1~~\\
     \bottomrule
    \end{tabular}
    \vspace{0.25cm}
    \caption{\textbf{Ablation on Window Sizes.} We conduct ablation study in 3DLoMatch and empirically found 11x11 window size shows the best performance. Two-view projection is used. SuperGlue~\cite{sarlin20superglue} is used to find correspondences and Pri3D~\cite{hou2021pri3d} pre-trained model is used for feature extraction.}
    \label{tab:window_size_ablation}
    \vspace{-0.5cm}
\end{table}

\begin{table}[h!]
    \renewcommand{\arraystretch}{1.2}
    \centering
    \begin{tabular}{l|c|ccccc}\toprule
&~\textit{Views}~   & \multicolumn{5}{c}{\textit{Registration Recall (\%)}} \\ 
         &     &~~5000~&~2500~&~1000~&~500~&~250~~\\ 
    \midrule
     \OURS{}~  &   1    &~~62.0~&~64.2~&~64.5~&~63.4~&~61.2~~\\
     \OURS{}~  &   2   &~~63.4~&~65.4~&~66.0~&~65.2~&~61.4~~\\
     \OURS{}~  &   3    &~~\textBF{63.7}~&~\textBF{65.5}~&~\textBF{66.1}~&~\textBF{65.2}~&~\textBF{61.7}~~\\
     \bottomrule
    \end{tabular}
    \vspace{0.25cm}
    \caption{\textbf{Ablation on Color Coverage.} In this table, we show a clear trend of increasing numbers on registration recall with more views used in the training and testing.}
    \label{tab:color_ablation}
    \vspace{-0.5cm}
\end{table}

\noindent \textbf{2D Pre-trained Networks.} \OURS{} lifts the deep color features into 3D. Therefore, massive existing 2D pre-trained models can be used. In Table.~\ref{tab:2d_backbone}, we show the influence of 2D models on 3D results. We ablate on popular 2D pre-trained weights, such as ImageNet~\cite{deng2009imagenet} and Pri3D~\cite{hou2021pri3d} pre-trained weights. With 2D pre-trained models, we can achieve better 3D results compared to random initializations (Scratch). In general, we can see a clear trend that our method can achieve better registration recall with powerful 2D pre-trained models. \\

\begin{table}[h!]
    \renewcommand{\arraystretch}{1.2}
    \centering
    \begin{tabular}{l|c|ccccc}
    \toprule
    & \textit{2D Backbone}   & \multicolumn{5}{c}{\textit{Registration Recall (\%)}} \\
      &     &~5000~&~2500~&~1000~&~500~&~250~\\ \midrule
    \OURS{}~ & Scratch  &~66.1~&~67.2~&~68.2~&~68.3~&~64.7~\\  
     \OURS{}~ &ImageNet &~66.3~&~\textBF{67.9}~&~68.9~&66.1~&~65.0~\\
     \OURS{}~ &Pri3D&~\textBF{66.3}~&~67.2~&~\textBF{69.0}~&~\textBF{68.5}~&~\textBF{65.0}~\\
     \bottomrule
    \end{tabular}
    \vspace{0.25cm}
    \caption{\textbf{Ablation on 2D Pre-trained Models.} We can see a correlation of 2D pre-trained weights and 3D performance when explicitly lifting deep color features. Using 2D pre-trained weights indicates higher registration recalls. Two-view projection is used. Note that all 2D models are not pre-trained on \textit{3DMatch} or \textit{3DLoMatch} data, thus showing a strong transfer ability in our proposed method.}
    \label{tab:2d_backbone}
    \vspace{-0.5cm}
\end{table}

\noindent \textbf{Implicit vs. Explicit Projection.} Adding color to 3D is non-trival. As aforementioned, we explore different ways of projecting 2D into 3D. Implicit one projects all the pixel values/features onto 3D, while explicit one leverages the 2D overlap information to project features region by region. We experimentally show that our design outperforms the rest. In Table.~\ref{tab:design}, we show different combinations of 2D pre-trained weights and projections. In general, projecting deep color features such as Pri3D outperforms SIFT features and RGB values. In addition, we show that explicit projection outperforms implicit projection \\

\begin{table*}[h]
    \centering
    \begin{tabular}{l|c|c|ccccc}
    \toprule
    \textit{Method}& ~\textit{Features}~ & ~\textit{Projection}~ & \multicolumn{5}{c}{\textit{Registration Recall (\%)}} \\
    &    & &~~5000~~&~~2500~~&~~1000~~&~~500~~&~~250~~\\ \midrule
     \OURS{}~  &  ~RGB~ & implicit &~~60.5~~&~~63.0~~&~~63.6~~&~~62.3~~&~~59.4~~\\
     \OURS{}~  &  ~RGB~ & explicit &~~60.4~~&~~63.1~~&~~63.5~~&~~62.8~~&~~59.9~~\\ 
      \OURS{}~ & ~SIFT~ & implicit &~~63.1~~&~~65.1~~&~~65.5~~&~~64.9~~&~~61.4~~\\  
      \OURS{}~ & ~SIFT~ & explicit &~~64.8~~&~~67.0~~&~~67.1~~&~~66.5~~&~~63.9~~\\
      \midrule
    \OURS{}~ &  ~SuperGlue~  & explicit &~~64.0~~&~~65.0~~&~~65.0~~&~~65.0~~&~~60.8~~\\
     \OURS{}~ &  ~ImageNet~  & implicit &~~63.2~~&~~65.4~~&~~65.7~~&~~64.9~~&~~61.1~~\\
      \OURS{}~ &  ~ImageNet~ & explicit &~66.3~&~\textBF{67.9}~&~68.9~&~66.1~&~65.0~~\\      
     \OURS{}~ &  ~Pri3D~  & implicit &~63.4~&~65.4~&~66.0~&~65.2~&~61.4~\\
     \OURS{}~ &  ~Pri3D~ & explicit&~\textBF{66.3}~&~67.2~&~\textBF{69.0}~&~\textBF{68.5}~&~\textBF{65.0}~\\
     \bottomrule
    \end{tabular}
    \vspace{.25cm}
    \caption{\textbf{Ablation on Projections.} RGB means simply appending RGB colors to point cloud. SIFT refers to projecting SIFT features onto points. Pri3D uses pre-trained weights to extract per-pixel features and projects them onto points. Similarly, SuperGlue/ImageNet refers to projecting SuperGlue/ImageNet pre-trained features. We show projecting deep color features outperforms SIFT and RGB values with the same projection. Implicit manner projects features of every pixel onto 3D, while explicit one projects features based on correspondences estimated by SuperGlue. We demonstrate the explicit projection surpasses the implicit one. Two-view projection is used in the experiments.}
    \label{tab:design}
\end{table*}

\noindent \textbf{Different Baselines.} Our approach is built upon Predator where our method adopts the same backbone and registration pipeline. However, our module is not specifically tied to Predator. Notably, our approach is agnostic to different SOTA methods in point cloud registration, and it is easy to be adapted into any frameworks operating on RGB-D data. In Table.~\ref{tab:cofinet}, we show our module also brings a significant improvement on CoFiNet baseline, i.e., +3.5\% Registration Recall. Similarly in Table~\ref{tab:geotransformer}, we observe an improvement of $2.4\%$ registration recall by incorporating our method onto GeoTransformer~\cite{qin2022geometric}.

\begin{table}[h!]
    \renewcommand{\arraystretch}{1.2}
    \centering
    \begin{tabular}{l|ccccc}
    \toprule
     &   \multicolumn{5}{c}{\textit{Registration Recall (\%)}} \\
  \textit{Baseline Method}~~~~~~~ &~~5000~~&~~2500~~&~~1000~~&~~500~~&~~250~~\\ \midrule
  CoFiNet    &~~67.5~&~66.2~&~64.2~&~63.1~&~61.0~~\\  
   CoFiNet (re-train)    &~~64.4~&~64.2~&~63.1~&~62.1~&~59.8~~\\ 
CoFiNet + \OURS{}   &~~\textBF{67.9}~&~\textBF{67.0}~&~\textBF{65.4}~& ~\textBF{64.2}~&~\textBF{62.2}~~\\  
     \bottomrule
    \end{tabular}
    \vspace{0.3cm}
    \caption{\textbf{Registration Recall based on CoFiNet in 3DLoMatch}. We can see a clear gap of plugging in our module compared to CoFiNet baseline. In this ablation experiment, our implementation is built upon the officially released code of CoFiNet. Thus, we re-train the official released code for a fair comparison. Pri3D pre-trained model and the two-view explicit projection is used}
    \label{tab:cofinet}
    \vspace{-0.5cm}
\end{table}

\begin{table}[h]
    \centering
    \begin{tabular}{l|cc}\toprule
         &\multicolumn{2}{c}{\textit{Registration Recall}} \\ 
            &~~~3DMatch~~~&~~~3DLoMatch~~~\\ 
            \midrule
            GeoTransformer~\cite{qin2022geometric}~ &   91.5 & 74.0 \\             
            GeoTransformer + \OURS{}~ &   92.5 & 76.4 \\
            \bottomrule
    \end{tabular}
    \vspace{0.25cm}
    \caption{\textbf{Registration Recall with GeoTransformer as Baseline.} Our method improves GeoTransformer's performance on both \textit{3DMatch} and \textit{3DLoMatch} benchmarks by noticeable margins. The experimental results further demonstrate that our approach is agnostic to backbones. Local to global registration is used to compute the registration recall.}
    \label{tab:geotransformer}
    \vspace{-0.5cm}
\end{table}
\section{Conclusion}
In this work, we focus on embedding color signals into geometry representations for point cloud registration. We notice that most of previous methods only use geometry as input, whereas RGB-D datasets naturally contain the RGB images. In this context, we propose our 2D-3D projection module to explicitly lift 2D features into 3D. We show that it is non-trivial to correlate color signals into geometry. Our experimental results show a significant improvement by using our projection module.  Our module is able to leverage massive existing work from 2D representation learning. Our approach is the first step towards geometry and color cross-modality learning in point cloud registration task, and there is a potential of making long-term impact with such cross-modality learning in many fields. We hope our research can inspire the community to pay more attention to this newly-developing direction.

\vspace{0.35cm}
\noindent \textbf{Acknowledgments} This work is supported by the National Key R\&D Program of China (2021YFC3320301) and the Joint Funds of Zhejiang NSFC (LTY22F020001) and Open Research Fund of State Key Laboratory of Transient Optics and Photonics.%

%
%
\bibliographystyle{splncs04}
\bibliography{egbib}

\clearpage
\noindent \textbf{\Large Appendix} \\
\begin{appendix}
In Section~\ref{sec:metric}, we introduce the details of the metrics used in the main paper. We additionally show more visual results in Section~\ref{sec:more}.

\begin{figure}[h!]
\begin{center}
\vspace{-0.55cm}
\includegraphics[width=0.9\linewidth]{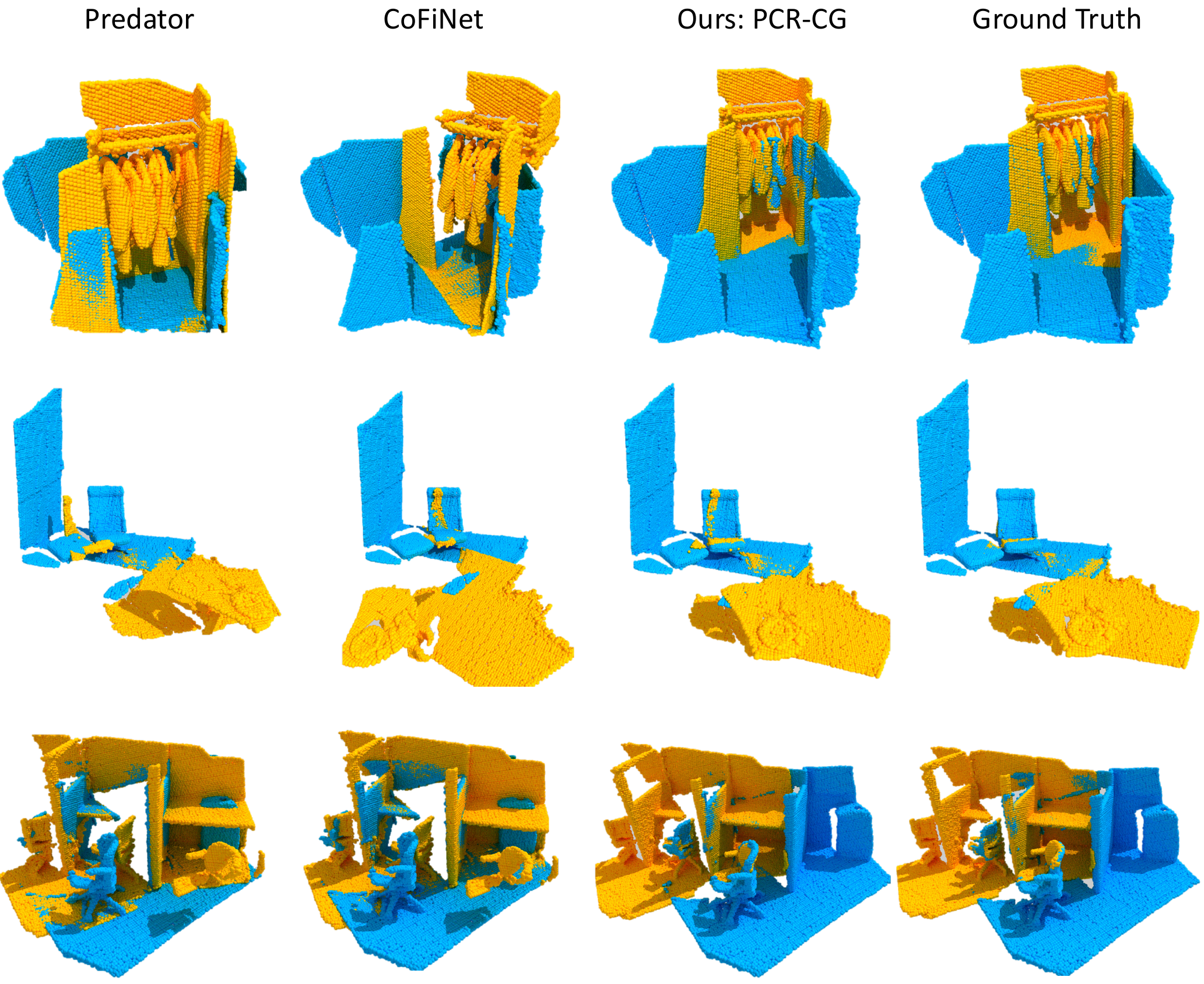}
\end{center}
\vspace{-0.55cm}
   \caption{\textbf{More Visualizations.} Our approach outperforms Predator~\cite{Huang_2021_CVPR} in most difficult scenarios in 3DLoMatch benchmark. }
   \vspace{-1.0cm}
\label{fig:more}
\end{figure}

\section{Evaluation Metrics}
\label{sec:metric}
\noindent \textbf{Inlier Ratio} (IR) measures the fraction of point correspondences $(x_{i}, y_{j}) \in \mathbf{\widetilde{\mathbf{C}}}$ subject to the Euclidean Norm of residual $\lVert \overline{\mathbf{T}}^{\mathbf{X}}_{\mathbf{Y}}(x_i) - y_j \rVert$. $\mathbf{\widetilde{\mathbf{C}}}$ denotes the estimated correspondence set.  $\overline{\mathbf{T}}^{\mathbf{X}}_{\mathbf{Y}}$ indicates the ground truth transformation between $\mathbf{X}$ and $\mathbf{Y}$. In the metric, we select threshold ${\tau}_1$=10cm. To this regard, a pair of correspondences count as matched when their Euclidean Norm of residual is smaller than 10cm. Given the  estimated correspondence set $\widetilde{\mathbf{C}}$, \textit{Inlier Ratio} of a pair of point clouds $(\mathbf{X}, \mathbf{Y})$ can be calculated by:

\begin{equation}
    \text{IR}(\mathbf{X}, \mathbf{Y}) = \frac{1}{\mid\widetilde{\mathbf{C}}\mid} \sum\limits_{(x_i, y_j)\in\widetilde{\mathbf{C}}} \mathbf{1} (\lVert \overline{\mathbf{T}}^{\mathbf{X}}_{\mathbf{Y}}(x_i) - y_j\rVert < {\tau}_1)
    \label{eq.ir}
\end{equation}

where $\mathbf{1}(\cdot)$ is the indicator function counting the number of correspondences within the threshold ${\tau}_1$; and $\lVert \cdot \rVert=\lVert \cdot \rVert_2$ denotes $L_2$ distance.\\

\noindent \textbf{Feature Matching Recall} (FMR) measures the fraction of point cloud pairs whose \textit{Inlier Ratio} is larger than a certain threshold $\tau_2=5\%$. It is firstly used in~\cite{deng2018ppf} and indicates the likelihood of recovering an optimal transformation between two point clouds by a robust pose estimator, e.g., RANSAC~\cite{fischler1981random}, based on the predicted correspondence set $\widetilde{\mathbf{C}}$. Given a dataset $\mathcal{D}$ with $|\mathcal{D}|$ point cloud pairs, \textit{Feature Matching Recall} can be computed as:

\begin{equation}
    \vspace{-0.1cm}
    \text{FMR}(\mathcal{D}) = \frac{1}{|\mathcal{D}|}\sum\limits_{(\mathbf{X}, \mathbf{Y})
    \in \mathcal{D}} \mathbf{1}(\text{IR}(\mathbf{X}, \mathbf{Y}) > \tau_2).
    \label{eq.fmr}
    \vspace{-0.1cm}
\end{equation}

\noindent \textbf{Registration Recall.} Different from aforementioned metrics that measure the quality of extracted correspondences, \textit{Registration Recall} (RR) on the other hand directly measures the performance on the task of point cloud registration. It measures the fraction of point cloud pairs whose Root Mean Square Error (RMSE) is within a certain threshold $\tau_3$ = 0.2m. Given a dataset $\mathcal{D}$ with $|\mathcal{D}|$ point cloud pairs, \textit{Registration Recall} is defined as:

\begin{equation}
    \text{RR}(\mathcal{D}) = \frac{1}{|\mathcal{D}|}\sum\limits_{(\mathbf{X}, \mathbf{Y})
    \in \mathcal{D}} \mathbf{1}(\text{RMSE}(\mathbf{X}, \mathbf{Y}) < \tau_3),
    \label{eq.rr}
\end{equation}
where for each $(\mathbf{X}, \mathbf{Y})\in \mathcal{D}$; RMSE is computed as following:

\begin{equation}
    \text{RMSE}(\mathbf{X}, \mathbf{Y}) = \sqrt{\frac{1}{|\overline{\mathbf{C}}|}\sum\limits_{(x_i, y_j)\in\overline{\mathbf{C}}}\lVert \mathbf{T}^{\mathbf{X}}_{\mathbf{Y}}(x_i) - y_j\rVert^{2}}.
    \label{eq.rmse}
\end{equation}
where the estimated transformation $\mathbf{T}^{\mathbf{X}}_{\mathbf{Y}}$ is applied. Additionally, we follow the original evaluation protocol in 3DMatch~\cite{zeng20173dmatch}, which excludes immediately adjacent point clouds with very high overlap ratios. \\

\noindent \textbf{Relative Translation and Rotation Errors.} Given the estimated transformation $\mathbf{T}^{\mathbf{X}}_{\mathbf{Y}} \in SE(3)$ composed of a translation vector $\mathbf{t} \in \mathbb{R}^{3}$ and a rotation matrix $\mathbf{R}\in SO(3)$. Its Relative Translation Error (RTE) and Relative Rotation Error (RRE) from the ground truth pose $\overline{\mathbf{T}}^{\mathbf{X}}_{\mathbf{Y}}$ are computed as:

\begin{equation}
\text{RTE} = \lVert\mathbf{t} - \overline{\mathbf{t}}\rVert \qquad \text{and} \qquad
\text{RRE} = \text{arccos}(\frac{\text{trace}(\mathbf{R}^{\top}\overline{\mathbf{R}}) - 1}{2}),
\label{eq.relative}
\end{equation}
where $\overline{\mathbf{t}}$ and $\overline{\mathbf{R}}$ are the the ground truth translation and rotation in $\overline{\mathbf{T}}^{\mathbf{X}}_{\mathbf{Y}}$, respectively.

\section{More Visualizations}
\label{sec:more}

We show more visualizations of our method compared to Predator~\cite{Huang_2021_CVPR} in Figure~\ref{fig:more}. Our approach can significantly improve registration accuracy in those more difficult scenarios.
\end{appendix}

\end{document}